\documentclass[10pt,twocolumn,letterpaper]{article}

\usepackage{cvpr}
\usepackage{times}
\usepackage{epsfig}
\usepackage{graphicx}
\usepackage{amsmath}
\usepackage{amssymb}
\usepackage{xcolor}
\usepackage{graphicx}
\usepackage{booktabs}
\usepackage{breqn}
\usepackage{caption}
\usepackage{subcaption}
\usepackage{dblfloatfix}
\usepackage{array}
\usepackage{balance}
\usepackage{titling}
\usepackage{xpatch}

\newcolumntype{x}[1]{>{\centering\arraybackslash\hspace{0pt}}p{#1}}

\usepackage[pagebackref=true,breaklinks=true,letterpaper=true,colorlinks,bookmarks=false]{hyperref}

\cvprfinalcopy 

\def\cvprPaperID{0914}

\newcommand{\PAR}[1]{\vskip3pt \noindent{\bf #1~}}
\newcommand{\circled}[1]{\raisebox{.5pt}{\textcircled{\raisebox{-.9pt} {#1}}}}

\makeatletter
\newenvironment{mytitlepage}
{\begin{titlepage}\def\@thanks{}}
	{\end{titlepage}}
\makeatother

\ifcvprfinal
\begin{document}
\pagenumbering{gobble}

\title{\textbf{A Neural Multi-sequence Alignment TeCHnique (NeuMATCH)}}

\author{Pelin Dogan\textsuperscript{1,4}\thanks{The technique was conceived when all authors worked for Disney Research.} \quad Boyang Li\textsuperscript{2} \quad Leonid Sigal\textsuperscript{3} \quad Markus Gross\textsuperscript{1,4}\\
	\textsuperscript{1}ETH Z{\"u}rich \quad \textsuperscript{2}Liulishuo AI Lab
	\quad \textsuperscript{3}University of British Columbia \quad \textsuperscript{4}Disney Research 
	\\%
	{\tt\small{\{pelin.dogan, grossm\}@inf.ethz.ch}, albert.li@liulishuo.com, lsigal@cs.ubc.ca}}

\date{\vspace{-2ex}}
\maketitle
\begin{abstract}
The alignment of heterogeneous sequential data (video to text) is an important and challenging problem. 
Standard techniques for this task, including Dynamic Time Warping (DTW) and Conditional Random Fields (CRFs), 
suffer from inherent drawbacks. Mainly, the Markov assumption implies that, given the immediate past, future 
alignment decisions are independent of further history. The separation between similarity computation and 
alignment decision also prevents end-to-end training. 
In this paper, we propose an end-to-end neural architecture where alignment actions are implemented as 
moving data between stacks of Long Short-term Memory (LSTM) blocks. This flexible architecture supports 
a large variety of alignment tasks, including one-to-one, one-to-many, skipping unmatched elements, and 
(with extensions) non-monotonic alignment. 
Extensive experiments on semi-synthetic and real datasets show that our algorithm outperforms 
state-of-the-art baselines.

\end{abstract}

\section{Introduction}

\begin{figure}[t]
    \centering
	\includegraphics[width=0.8\columnwidth]{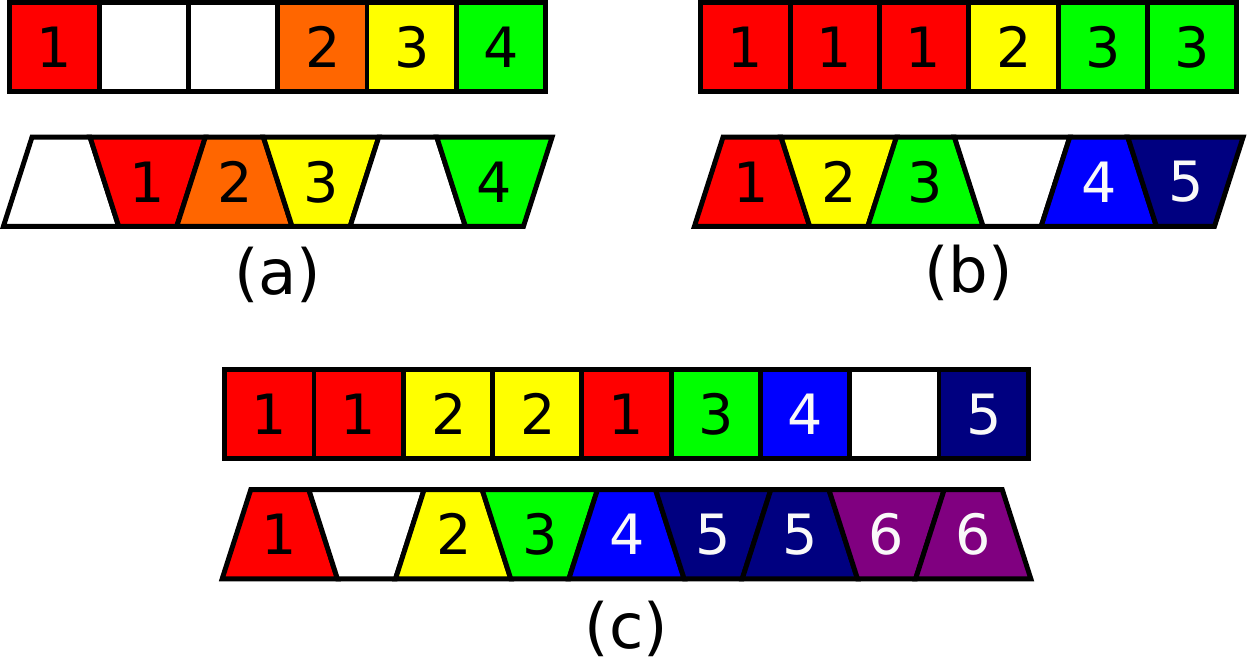}
	\caption{Types of sequence correspondence. Matching blocks in two sequences have identical colors and numbers. (a) A one-to-one matching where the white blocks do not match anything. (b) A one-to-many matching where one block on the bottom sequence matches multiple blocks on the top. (c) A non-monotonic situation where the matching does not always proceed strictly from left to right due to the red-1 block after the yellow-2 on top. }
        \label{fig:alig}
	\vspace{-12pt}
\end{figure}

Sequence alignment (see Figure \ref{fig:alig}) is a prevalent problem that finds diverse applications in molecular biology \cite{Loytynoja2005}, natural language processing \cite{barzilay2003learning}, historic linguistics \cite{prokic2009multiple}, and computer vision \cite{caspi2000step}. 
In this paper, we focus on aligning heterogeneous sequences 
with complex correspondences. 
Heterogeneity refers to the lack of an obvious surface matching (a literal similarity metric between elements of the sequences). A prime example is the 
alignment between visual and textual content. Such alignment 
requires sophisticated extraction of comparable feature representations in each modality, often performed by a deep neural network. 

A common solution to the alignment problem consists of two stages that are performed separately: (1) the learning of a similarity metric between elements in the sequences and (2) finding the optimal alignment between the sequences. 
Alignment techniques based on dynamic programming, such as Dynamic Time Warping (DTW)~\cite{berndt1994using} and Canonical Time Warping (CTW)~\cite{ZhouD16}, are widely popular.
In a simple form, DTW can be understood as finding the shortest path where the edge costs are computed with the similarity metric, so the decision is Markov. Variations of DTW \cite{tapaswi2015book2movie,zhu2015aligning} accommodate some degrees of non-monotonicity (see Figure \ref{fig:alig} (c)). 
In all cases, these approaches are disadvantaged by the separation of the two stages. Conceptually, learning a metric that directly helps to optimize alignment should be beneficial. Further, methods with first-order Markov assumptions take only limited local context into account, but contextual information conducive to alignment may be scattered over the entire sequence. For example, knowledge of the narrative structure of a movie may help to align shots to their sentence descriptions. 

\begin{figure*}
	\vspace{-7pt}
	\centering
	\includegraphics[width=\textwidth]{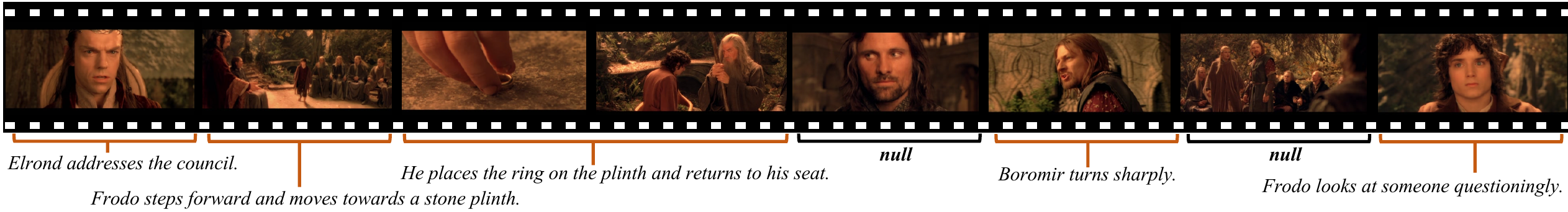}
	\label{fig:lotr}
	\vspace{-15pt}
	\caption{An example alignment between clip sequence and text sequence (from the dataset HM-2 in Section \ref{subsec:data}). 
	}
	\label{fig:lotr}
	\vspace{-12pt}
\end{figure*}

To address these limitations, we propose an end-to-end differentiable neural architecture for heterogeneous sequence alignment, which we call NeuMATCH.
The NeuMATCH architecture represents the current state of the workspace using four Long Short-term Memory (LSTM) chains: two for the partially aligned sequences, one for the matched content, and one for historical alignment decisions. The four recurrent LSTM networks collectively capture the decision context, which is then classified into one of the available alignment actions. 
Compared to the traditional two-stage solution, the network can be optimized end-to-end. In addition, the previously matched content and the decision history inform future alignment decisions in a non-Markov manner. For example, if we match a person's face with the name Frodo at the beginning of a movie, we should be able to identify the same person again later (Figure \ref{fig:lotr}). Alternatively, if the input sequences are sampled at different rates (e.g., every third video clip is matched to text), 
the decision history can help to discover and exploit such regularities.

Although the proposed framework can be applied
to different types of sequential data, in this paper, we focus on the alignment of video and textual sequences, especially those containing narrative content like movies. This task is an important link in joint understanding of multimodal content \cite{fang2015captions} and is closely related to activity recognition \cite{deng2016structure,wang2011action}, dense caption generation \cite{krishna2017dense}, and multimedia content retrieval \cite{kiros2014unifying,vendrov2015order}. The reason for choosing narrative content is that it is among the most challenging for computational understanding due to a multitude of causal and temporal interactions between events~\cite{sheinfeld2016video}.  Disambiguation is difficult with needed contextual information 
positioned far apart. Thus, narrative contents make an ideal application and testbed for alignment algorithms. 

\vspace{0.1in}
\noindent
{\bf Contributions. }
The contributions of this paper are two-fold. First, we propose a novel end-to-end neural framework for heterogeneous multi-sequence alignment. Unlike prior methods, our architecture is able to take into account rich context when making alignment decisions.
Extensive experiments illustrate that the framework {\em significantly} outperforms traditional baselines in accuracy. Second, we annotate a new dataset\footnote{https://github.com/pelindogan/NeuMATCH} containing movie summary videos and share it with the research community.

 \section{Related Work}

\begin{table*}
\vspace{-7pt}
\centering
\small
\begin{tabular}{lcx{3cm}cccc}
\toprule
& \cite{sankar2009subtitle} & \cite{zhu2015aligning} & \cite{tapaswi2015book2movie} & \cite{tapaswi2014story} & \cite{bojanowski2015weakly} &  {\bf NeuMATCH} \\ \midrule
\textbf{Method} & DTW & CRF Chain & DP & DP & QIP & Neural \\
\textbf{End-to-end Training} & No & No & No & No & No & Yes \\
\textbf{Historic Context} & Markov & Markov + Convolution on Similarity  & Markov & Markov & global & high order \\
\textbf{Supports Non-monotonicity} & No & Yes & Yes & No & No & Yes* \\
\textbf{Visual/Textual Granularity} & fine & medium & coarse & fine & fine & fine \\
\bottomrule
\end{tabular}
\vspace{-5pt}
\caption{ Comparison of existing video-text alignment approaches. Prior method are based on DTW/Dynamic Programming (DP), Conditional Random Field (CRF) and Convex Quadratic Programming (CQP). *Non-monotonicity requires extensions in Appendix~\ref{sec:extensions}.}
\label{tab:related}
\vspace{-10pt}
\end{table*}

Our goal of video-text alignment is related to multiple topics. 
We briefly review the most relevant literature below. 

\PAR{Unimodal Representations.}
It has been observed that deep convolutional neural networks (CNNs), such as VGG~\cite{simonyan2014very}, ResNet~\cite{he2016deep}, GoogLeNet~\cite{szegedy2015going}, and even automatically learned architectures~\cite{zoph2017}, can learn image features that are transferable to many different vision tasks \cite{Donahue2014,Yosinski:2014}. 
Generic representations for video and text have received comparatively less attention. Common encoding techniques for video include pooling~\cite{venugopalan2014translating} and attention \cite{xu2015show,yao2015describing} over frame features, neural recurrence between frames~\cite{donahue2015long, ranzato2014video,venugopalan2015sequence}, and  
spatiotemporal 3D convolution~\cite{tran2015learning}. 
On the language side, distributed word representations \cite{mikolov2013efficient,pennington2014glove} are often used in recurrent architectures in order to model sentential semantics. When coupled with carefully designed training objectives, such as Skip-Thought~\cite{kiros2015skip} or textual entailment~\cite{Bowman2015,conneau2017}, they yield effective
representations that generalize well to other tasks. 

\PAR{Joint Reasoning of Video and Text.}
Popular research topics in joint reasoning and understanding of visual and textual information include image captioning \cite{karpathy2015deep, mao2014deep, vinyals2015show, xu2015show}, retrieval of visual content~\cite{lin2014visual}, and visual question answering \cite{antol2015vqa, sadeghi2015viske, xu2016ask}.
Most approaches along these lines can be classified as belonging to either (i) joint language-visual embeddings or (ii) encoder-decoder architectures. The joint {\em vision-language embeddings} facilitate image/video or caption/sentence retrieval by learning to embed images/videos and sentences into the same space \cite{pan2016jointly, torabi2016learning, Xu2017Emotion, xu2015jointly}. For example, \cite{hodosh2013framing} uses simple kernel CCA and in \cite{farhadi2010every} both images and sentences are mapped into a common semantic {\em meaning} space defined by object-action-scene triplets. More recent methods directly minimize a pairwise ranking function between positive image-caption pairs and contrastive (non-descriptive) negative pairs; various ranking objective functions have been proposed including max-margin \cite{kiros2014unifying} and order-preserving losses \cite{vendrov2015order}. 
The {\em encoder-decoder} architectures \cite{torabi2016learning} are similar, but instead attempt to encode images into the embedding space from which a sentence can be decoded. Applications of these approaches for video captioning and dense video captioning (multiple sentences) were explored in \cite{pan2016jointly} and \cite{yu2016video} respectively, for video retrieval in \cite{donahue2015long}, and for visual question answering in \cite{Anderson2017}.
In this work, we jointly encode the video and textual input as part of the decision context. Instead of decoding alignment decisions one by one with RNNs,  we gather the most relevant contexts for every alignment decision and directly predict the decision from those.

\PAR{Video-text alignment.}
Under the dynamic time warping framework, early works on video/image-text alignment adopted a feature-rich approach, utilizing features from dialogs and subtitles~\cite{cour2008movie, everingham2006hello, tapaswi2014story}, location, face and speech recognition~\cite{sankar2009subtitle}, as well as nouns and pronouns between text and objects in the scenes~\cite{dogan2016label, kong2014you, lin2014visual, malmaud2015s}. 

Tapaswi \emph{et al.} \cite{tapaswi2014story} present an approach to align plot synopses with the corresponding shots with the guidance of subtitles and facial features from characters. They extend the DTW algorithm to allow one-to-many matching.
In~\cite{tapaswi2015book2movie}, Tapaswi \etal present another extension to allow non-monotonic matching in the alignment of book chapters and video scenes. The above formulations make use of the Markov property, which enables efficient solutions with dynamic programming (DP). At the same time, the historic context being considered is limited.
\cite{zhu2015aligning} develops neural approach for the computation of similarities between videos and book chapters, using Skip-Thought vectors \cite{kiros2015skip}. In order to capture historic context, they use a convolutional network over a similarity tensor. The alignment is formulated as a linear-chain Conditional Random Field (CRF), which again yields efficient solution from DP. Although this method considers historic context, the alignment and similarity are still computed separately. 

Bojanowski \etal~\cite{bojanowski2015weakly} formulate alignment as quadratic integer programming (QIP) and solve the relaxed problem. 
Weak supervision can be introduced as optimization constraints. 
This method considers the global context, but relates the video and text features by a linear transformation and does not consider non-monotonic alignment. Table \ref{tab:related} compares key aspects of these methods.

In summary, existing approaches perform the alignment in two separate stages: (1) extracting visual and textual features in such a way as to have a well defined metric, and (2) performing the alignment using this similarity (and possibly additional side information). We propose an end-to-end differentiable neural architecture that considers more than the local similarities. Inspired by LSTM-powered shift-reduce language parsers~\cite{dyer2015transition,honnibal2015improved}, we augment LSTM networks with stack operations, such as pop and push. The advantage of this setup is that the most relevant video clips, sentences, and historic records are always positioned closest to the prediction. 
 \section{Approach}

\begin{figure*}[t]
    \centering
	\includegraphics[width=2\columnwidth]{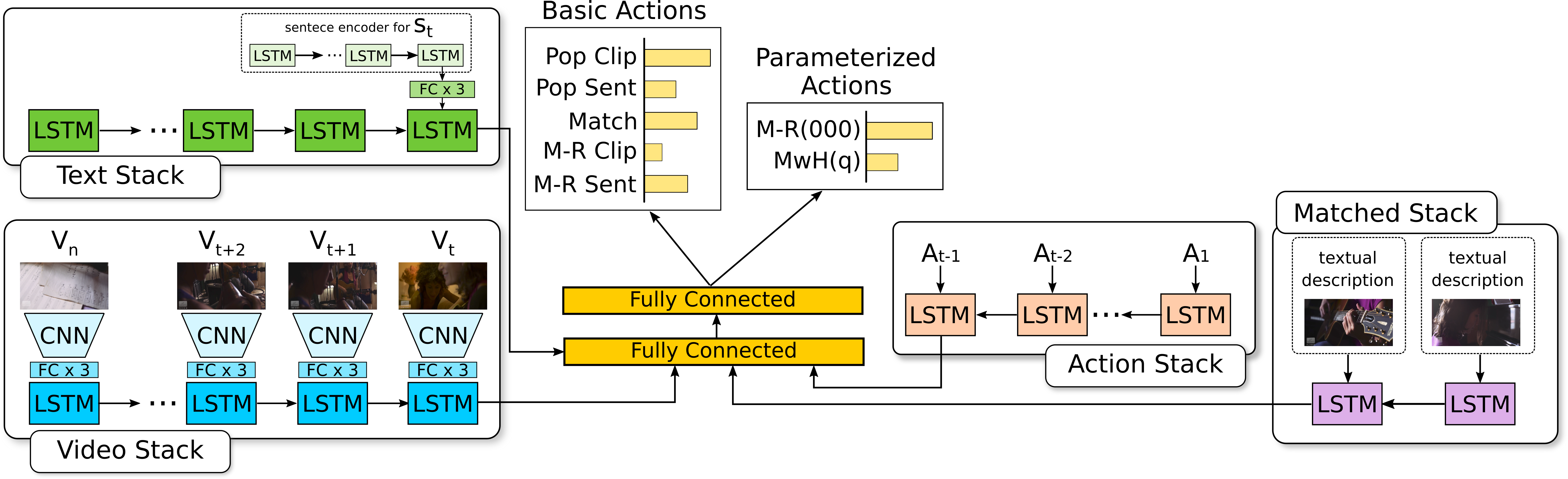}
	\vspace{-5pt}
	\caption{The proposed NeuMATCH neural architecture. The current state as described by the four LSTM chains is classified into one of the alignment decisions. Parameterized actions are explained and illustrated in Appendix~\ref{sec:extensions} and Table~\ref{tab:multi-seq-actions}.}
	\label{fig:stackArchitecture}
	\vspace{-4pt}
\end{figure*}
We now present NeuMATCH, a neural architecture for temporal alignment of heterogeneous sequences.  
While the network is general, for this paper we focus specifically on the 
video and textual sequence alignment.
The video sequence consists of a number of consecutive video clips $\mathcal{V} = \left\{V_i\right\}_{i=1\ldots N}$. 
The textual sequence consists a number of consecutive sentences $\mathcal{S} = \left\{S_i\right\}_{i=1\ldots M}$.
Our task is to align these two sequences by, for example, finding a function $\pi$ that maps an index of the video 
segment to the corresponding sentence: $\langle V_i, S_{\pi(i)} \rangle$.
An example input for our algorithm can be a movie segmented into individual shots and the accompanying movie 
script describing the scenes and actions, which are broken down into sentences (Figure \ref{fig:lotr}). 
The video segmentation could be achieved using any shot boundary detection algorithm; NeuMATCH can handle one-to-many matching caused by over-segmentation. 

We observe that the most difficult sequence alignment problems exhibit the following characteristics. 
First, heterogeneous surface forms, such as video and text, can conceal the true similarity structure, which suggests 
a satisfactory understanding of the entire content may be necessary for alignment. Second, difficult problems contain 
complex correspondence like many-to-one matching and unmatched content, which the framework should accommodate. 
Third, contextual information that are needed for learning the similarity metric are scattered over the entire sequence. 
Thus, it is important to consider the history and the future when making the alignment decision and to create an 
end-to-end network where gradient from alignment decisions can inform content understanding and similarity metric learning. 

The NeuMATCH framework copes with these challenges by explicitly representing the state of the entire workspace, including the partially matched input sequences and historic alignment decisions. The representation employs four LSTM recurrent networks, including the input video sequence (Video Stack), the input textual sequence (Text Stack), previous alignment actions (Action Stack) as well as previous alignments themselves (Matched Stack). Figure \ref{fig:stackArchitecture} shows the NeuMATCH architecture.

We learn a function that maps the state of workspace $\Psi_t$ to an alignment action $A_t$ at every time step $t$. The action $A_t$ manipulates the content of the LSTM networks, resulting in a new state $\Psi_{t+1}$. Executing a complete sequence of actions produces an alignment of the input. The reader may recognize the similarity with policy gradient methods~\cite{Sutton2017}. 
As the correct action sequence is unique in most cases and can be easily inferred from the ground-truth labels, in this paper, we adopt a supervised learning approach. 

The alignment actions may be seen as stack operations because they either remove or insert an element at the first position of the LSTM network (except for non-monotonic matching discussed in Appendix~\ref{sec:extensions}).
For example, elements at the first position can be removed ({\em popped}) or {\em matched}. 
When two elements are matched, they are removed from the input stacks and stored in the Matched Stack.

\subsection{Language and Visual Encoders}
\label{subsec:encoder}

We first create encoders for each video clip and each sentence. After that, we perform an optional pre-training step to jointly embed the encoded video clips and sentences into the same space. While the pre-training step produces a good initialization, the entire framework is trained end-to-end, which allows the similarity metric to be specifically optimized for the alignment task.  

\PAR{Video Encoder.} 
We extract features using the activation of the first fully connected layer in the VGG-16 network \cite{simonyan2014very}, which produces a 4096-dim vector per frame. As each clip is relatively short and homogeneous, we perform mean pooling over all frames in the video, yielding a feature vector for the entire clip. This vector is transformed with three fully connected layers using the ReLU activation function, resulting in 
encoded video vector $v_i$ for the $i^{\text{th}}$ clip.

\PAR{Sentence Encoder.} 
The input text is parsed into sentences $S_1 \ldots S_M$, each of which contains a sequence of words. We transform each unique word into an embedding vector pre-trained using GloVe \cite{pennington2014glove}. The entire sentence is then encoded using a 2-layer LSTM recurrent network, where the hidden state of the first layer, $h^{(1)}_t$, is fed to the second layer:
\begin{subequations}
\begin{align}
h^{(1)}_t, c^{(1)}_t &= \text{LSTM}(x_t, h^{(1)}_{t-1}, c^{(1)}_{t-1}) \\
h^{(2)}_t, c^{(2)}_t &= \text{LSTM}(h^{(1)}_t, h^{(2)}_{t-1}, c^{(2)}_{t-1})
\enspace ,
\end{align}
\end{subequations}
where $c^{(1)}_t$ and $c^{(2)}_t$ are the memory cells for the two layers, respectively; $x_t$ is the word embedding for time step $t$. 
The sentence is represented as the vector obtained by the transformation of the last hidden state $h^{(2)}_t$ by three fully connected layers using ReLU activation function.

\PAR{Encoding Alignment and Pre-training.}
Due to the complexity of the video and textual encoders, we opt for pre-training that produces a good initialization for subsequent end-to-end training. 
For a ground-truth pair $(V_i, S_i)$, we adopt an asymmetric similarity proposed by \cite{vendrov2015order}
\begin{equation}
\label{eq:similarity}
F(v_i,s_i) = - ||\max(0, v_i-s_i)||^2
\enspace .
\end{equation}
This similarity function takes the maximum value 0, when $s_i$ is positioned to the upper right of $v_i$ in the vector space. That is, $\forall j, s_{i,j} \ge v_{i,j}$. When that condition is not satisfied, the similarity decreases. In \cite{vendrov2015order}, this relative spatial position defines an entailment relation where $v_i$ entails $s_i$. Here the intuition is that the video typically contains more information than being described in the text, so we may consider the text as entailed by the video.  

We adopt the following ranking loss objective by randomly sampling a contrastive video clip $V^\prime$ and a contrastive sentence $S^\prime$ for every ground truth pair. Minimizing the loss function maintains that the similarity of the contrastive pair is below true pair by at least the margin $\alpha$.  
\begin{dmath}
\label{eq:pairwiseLoss}
\mathcal{L} = \sum_{i}\left(\mathbb{E}_{v^\prime \neq v_i} \max \left\{0, \alpha-F(v_i, s_i)+F(v^\prime, s_i)\right\}  \\
		+ \mathbb{E}_{s^\prime \neq s_i}  \max\left\{0, \alpha-F(v_i, s_i)+F(v_i, s^\prime)\right\} \right)
\end{dmath}
Note the expectations are approximated by sampling. 

\subsection{The NeuMATCH Alignment Network}
\label{subsec:mach}

With the similarity metric between video and text acquired by pre-training, a naive approach for alignment is to maximize the collective similarity over the matched video clips and sentences. 
However, doing so ignores the temporal structures of the two sequences and can lead to degraded performance. NeuMATCH considers the history and the future by encoding input sequences and decision history with LSTM networks.

\PAR{LSTM Stacks.}
At time step $t$, the first stack contains the sequence of video clips yet to be processed $V_t, V_{t+1}, \ldots, V_{N}$. The direction of the LSTM goes from $V_N$ to $V_t$, which allows the information to flow from the future clips to the current clip.
We refer to this LSTM network as the video stack and denote its hidden state as $h^V_t$. Similarly, the text stack contains the sentence sequence yet to be processed: $S_t, S_{t+1}, \ldots, S_{M}$. Its hidden state is $h^S_t$. 

The third stack is the action stack, which stores all alignment actions performed in the past. The actions are denoted as $A_{t-1}, \ldots, A_1$ and are encoded as one-hot vectors $a_{t-1}, \ldots, a_1$. The reason for including this stack is to capture patterns in the historic actions. Different from the first two stacks, the information flows from the first action to the immediate past with the last hidden state being $h^A_{t-1}$.

The fourth stack is the matched stack, which contains only the texts and clips that are matched previously and places the last matched content at the top of the stack. We denote this sequence as $R_1, \ldots, R_L$. Similar to the action stack, the information flows from the past to the present. In this paper, we consider the case where one sentence $s_i$ can match with multiple video clips $v_1, \ldots, v_K$. Since the matched video clips are probably similar in content, we perform mean pooling over the video features $v_i = \sum_j^K v_j / K$. The input to the LSTM unit is hence the concatenation of the two modalities $r_i = [s_i, v_i]$. The last hidden state of the matched stack is $h^M_{t-1}$. 

\PAR{Alignment Action Prediction.}
At every time step, the state of the four stacks is $\Psi_t = (V_{t^+}, S_{t^+}, A_{(t-1)^-}, R_{1^+})$, where we use the shorthand $X_{t^+}$ for the sequence $X_t, X_{t+1}, \ldots$ and similarly for $X_{t^-}$. $\Psi_t$ can be approximately represented by the LSTM hidden states. Thus, the conditional probability of alignment action $A_t$ at time $t$ is
\begin{equation}
P(A_t | \Psi_t ) = P(A_t | h^V_t , h^S_t, h^A_{t-1}, h^M_{t-1})
\end{equation}
The above computation is implemented as a softmax operation after two fully connected layers with ReLU activation on top of the concatenated state $\psi_t = [h^V_t, h^S_t, h^A_{t-1}, h^M_{t-1}]$. In order to compute the alignment of entire sequences, we apply the chain rule.
\begin{equation}
P(A_1, \ldots, A_N | \mathcal{V}, \mathcal{S} ) = \prod_{t=1}^N P(A_t | A_{(t-1)^-}, \Psi_t )
\end{equation}
The probability can be optimized greedily by always choosing the most probable action or using beam search. 
The classification is trained in a supervised manner. From a ground truth alignment of two sequences, we can easily derive a correct sequence of actions, which are used in training. In the infrequent case when more than one correct action sequence exist, one is randomly picked. The training objective is to minimize the cross-entropy loss at every time step. 

\PAR{Alignment Actions.}
We propose five basic alignment actions that together handle the alignment of two sequences with unmatched elements and one-to-many matching. The actions include \emph{Pop Clip} (PC), \emph{Pop Sentence} (PS), \emph{Match} (M), \emph{Match-Retain Clip} (MRC), and \emph{Match-Retain Sentence} (MRS). Table \ref{tab:actions} provides a summary of their effects. 

\begin{table}[t]
	\centering
	\small
	\begin{tabular}{lp{3.1em}p{3em}p{3.7em}p{3.2em}}
		\toprule
		& \textbf{Video Stack} & \textbf{Text Stack} & \textbf{Matched Stack} & \textbf{Action Stack} \\
		\midrule
		\textbf{Initial} & \textcircled{a}\circled{b}\textcircled{c}  & \circled{1}\circled{2}\circled{3} &  & \\
		\midrule
		Pop Clip & \circled{b}\textcircled{c} & \circled{1}\circled{2}\circled{3} &  & PC \\
		Pop Sent & \textcircled{a}\circled{b}\textcircled{c} & \circled{2}\circled{3} &  & PS  \\
		Match & \circled{b}\textcircled{c} & \circled{2}\circled{3} & [\textcircled{a}\circled{1}] & M  \\		
		\midrule
		Match-Retain-C & \textcircled{a}\circled{b}\textcircled{c}  & \circled{2}\circled{3} & [\textcircled{a}\circled{1}] & MRC  \\
		Match-Retain-S & \circled{b}\textcircled{c}  & \circled{1}\circled{2}\circled{3} & [\textcircled{a}\circled{1}] & MRS   \\
		\bottomrule
	\end{tabular}
	\vspace{-5pt}
	\caption{The basic action inventory and their effects on the stacks. Square brackets indicate matched elements.}
	\label{tab:actions}
	\vspace{-7pt}
\end{table}

The Pop Clip action removes the top element, $V_t$, from the video stack. This is desirable when $V_t$ does not match any element in the text stack. Analogously, the \textit{Pop Sentence} action removes the top element in the text stack, $S_t$. 
The Match action removes both $V_t$ and $S_t$, matches them, and pushes them to the matched stack. 
The actions Match-Retain Clip and Match-Retain Sentence are only used for one-to-many correspondence. When many sentences can be matched with one video clip, the Match-Retain Clip action pops $S_t$, matches it with $V_t$ and pushes the pair to the matched stack, but $V_t$ stays on the video stack for the next possible sentence. To pop $V_t$, the Pop Clip action must be used. The Match-Retain Sentence action is similarly defined. In this formulation, matching is always between elements at the top of the stacks. 

It is worth noting that the five actions do not have to be used together. A subset can be picked based on knowledge about the sequences being matched. For example, for one-to-one matching, if we know some clips may not match any sentences, but every sentence have at least one matching clip, we only need Pop Clip and Match. Alternatively, consider a one-to-many scenario where (1) one sentence can match multiple video clips, (2) some clips are unmatched, and (3) every sentence has at least one matching clip. We need only the subset Pop Clip, Pop Sentence, and Match-Retain Sentence. It is desirable to choose as few actions as possible, because it simplifies training and reduces the branching factor during inference.  

\PAR{Discussion.}
The utility of the action stack becomes apparent in the one-to-many setting. As discussed earlier, to encode an element $R_i$ in the matched stack, features from different video clips are mean-pooled. As a result, if the algorithm needs to learn a constraint on how many clips can be merged together, features from the matched stack may not be effective, but features from action stack would carry the necessary information.
The alignment actions discussed in the above section allow monotonic matching for two sequences, which is the focus of this paper and experiments. We discuss extensions that allow multi-sequence matching as well as non-monotonic matching in Appendix~\ref{sec:extensions}.

 \section{Experimental Evaluation}
We evaluate NeuMATCH on semi-synthetic and real datasets, including a newly annotated, real-world YouTube Movie Summaries (YMS) dataset. Table~\ref{tab:statistics} shows the statistics of the datasets used. 

\subsection{Datasets}
\label{subsec:data}

We create the datasets HM-1 and HM-2 based on the LSMDC data~\cite{rohrbach2015dataset}, which contain matched clip-sentence pairs. The LSMDC data contain movie clips and very accurate textual descriptions, which are originally intended for the visually impaired. We generate video and textual sequences in the following way: First, video clips and their descriptions in the same movie are collected sequentially, creating the initial video and text sequences.
For HM-1, we randomly insert video clips from other movies into each video sequence. In order to increase the difficulty of alignment and to make the dataset more realistic, we select confounding clips that are similar to the neighboring clips. After randomly choosing an insertion position, we sample 10 video clips and select the most similar to its neighboring clips, using the pre-trained similarity metric (Section \ref{subsec:encoder}). An insertion position can be 0-3 clips away from the last insertion.
For HM-2, we randomly delete sentences from the collected text sequences. A deletion position is 0-3 sentences from the last deletion.
At this point, HM-1 and HM-2 does not require one-to-many matching, which is used to test the 2-action NeuMATCH model. To allow one-to-many matching, we further randomly split every video clip into 1-5 smaller clips. 

\PAR{YMS dataset.} We create the YMS dataset from the YouTube channels \textit{Movie Spoiler Alert} and \textit{Movies in Minutes}, where a narrator orally summarizes movies alongside clips from the actual movie. Two annotators transcribed the audio and aligned the narration text with video clips. The YMS dataset is the most challenging for several reasons: (1) The sequences are long. On average, a video sequence contains 161.5 clips and a textual sequence contains 58.2 sentences. (2) A sentence can match a long sequence of (up to 45) video clips. (3) Unlike LSMDC, YMS contains rich textual descriptions that are intended for storytelling; they are not always faithful descriptions of the video, which makes YMS a challenging benchmark. 

\begin{table}
\centering
\small
\begin{tabular}{lrrrr}
\toprule
           & \textbf{HM-1} & \textbf{HM-2} & \textbf{YMS} \\ \midrule
\# words & 4,196,633 & 4,198,021 &  54,326 \\
\# sent. & 458,557 & 458,830 & 5,470 \\
\# avg. words/sent. & 9.2 & 9.1 & 9.5 \\ \midrule
\# clips & 1,788,056 & 1,788,056 & 15,183 \\
\# video & 22,945 & 22,931 & 94 \\
\# avg clips/video & 77.9 & 77.9 & 161.5 \\ \midrule
\# avg sent./video & 20.0 & 20.0 & 58.2\\
\# clip/sent. (mean(var)) & 2.0(0.33) & 2.0(0.33) & 2.6(8.8) \\
\bottomrule
\end{tabular}
\vspace{-4pt}
\caption{Summary statistics of the datasets.}
\label{tab:statistics}
\vspace{-7pt}
\end{table}

\subsection{Performance Metrics}

For one-to-one matching, we measure the matching accuracy, or the percentage of sentences and video clips that are correctly matched or correctly assigned to {\em null}. For one-to-many matching, where one sentence can match multiple clips, we cannot use the same accuracy for sentences. Instead, we turn to the Jaccard Index, which measures the overlap between the predicted range and the ground truth of video clips using the intersection over union (IoU).

\begin{table}
	\centering
	\resizebox{\columnwidth}{!}{
		\begin{tabular}{lrrrrrrrr}
			\toprule
			& \multicolumn{4}{c}{\textbf{HM-1}} &\multicolumn{4}{c}{\textbf{HM-2}} \\		
			\cmidrule(lr){2-5}
			\cmidrule(lr){6-9}
			& \textit{MD} & \textit{CTW} & \textit{DTW} & \textit{\textbf{Ours}} & \textit{MD} & \textit{CTW} & \textit{DTW} & \textit{\textbf{Ours}} \\
			\midrule
			\textbf{clips} & 6.4 & 13.4 & 13.3 & \textbf{69.7} & 2.5 & 12.9 & 13.0 & \textbf{40.6} \\
			\textbf{sents.} & 15.8 & 21.3 & 41.7 & \textbf{58.6} & 15.6 & 25.1 & 34.2 & \textbf{43.7} \\
			\bottomrule
			\bottomrule
		\end{tabular}
	}
	\vspace{-3pt}
	\caption{Accuracy of clips and sentences for the 2-action model. Datasets require the detection of {\em null} clips.}
	\label{tab:results_action2}
	\vspace{-7pt}
\end{table}

\subsection{Baselines}
We create three baselines, Minimum Distance (MD), Dynamic Time Warping (DTW), and Canonical Time Warping (CTW). 
All baselines use the same jointly trained language-visual neural network encoders (Section~\ref{subsec:encoder}), which are carefully trained and exhibit strong performance. Due to space constraints, we discuss implementation details in the supplementary material. 

\begin{table*}
	\vspace{-5pt}
	\centering
	\resizebox{\textwidth}{!}{		
		\begin{tabular}{lcccc@{\hspace{10pt}}cccc@{\hspace{10pt}}cccc@{\hspace{10pt}}cccc}
			\toprule
			& \multicolumn{4}{c}{\textbf{HM-0}} & \multicolumn{4}{c}{\textbf{HM-1}} & 
			\multicolumn{4}{c}{\textbf{HM-2}} & \multicolumn{4}{c}{\textbf{YMS}} \\
			\cmidrule(lr){2-5}
			\cmidrule(lr){6-9}
			\cmidrule(lr){10-13}
			\cmidrule(lr){14-17}
			& \textit{MD} & \textit{CTW} & \textit{DTW} & \textit{\textbf{Ours}} &  \textit{MD} & \textit{CTW} & \textit{DTW} & \textit{\textbf{Ours}} &  \textit{MD} & \textit{CTW} & \textit{DTW} & \textit{\textbf{Ours}} & \textit{MD} & \textit{CTW} & \textit{DTW} & \textit{\textbf{Ours}} \\
			\midrule
			\textbf{clips} & 20.7 & 26.3 & 50.6 & \textbf{63.1} & 10.5 & 6.8 & 17.6 & \textbf{65.0} & 10.6 & 6.9 & 18.0 & \textbf{37.7} & 4.0 & 5.0 & 10.3 & \textbf{12.0} \\
			\textbf{sents IoU} & 23.0 & 25.4 & 42.8 & \textbf{55.3} & 5.7 & 7.3 & 18.4 & \textbf{44.1} & 9.0 & 7.6 & 18.9 & \textbf{20.0} & 2.4 & 3.6 & 7.5 & \textbf{10.4}\\
			\bottomrule
			\bottomrule
		\end{tabular}
	}
	\vspace{-5pt}
	\caption{Alignment performance for 3-action model given in percentage (\%) over all data. Datasets HM-1, HM-2, and YMS require the detection of null clips and one-to-many matchings of the sentences. HM-0 only requires one-to-many matching of sentences.}
	\label{tab:results_action3}
	\vspace{-7pt}
\end{table*}

\begin{figure*}
	\centering
	\includegraphics[width=\textwidth]{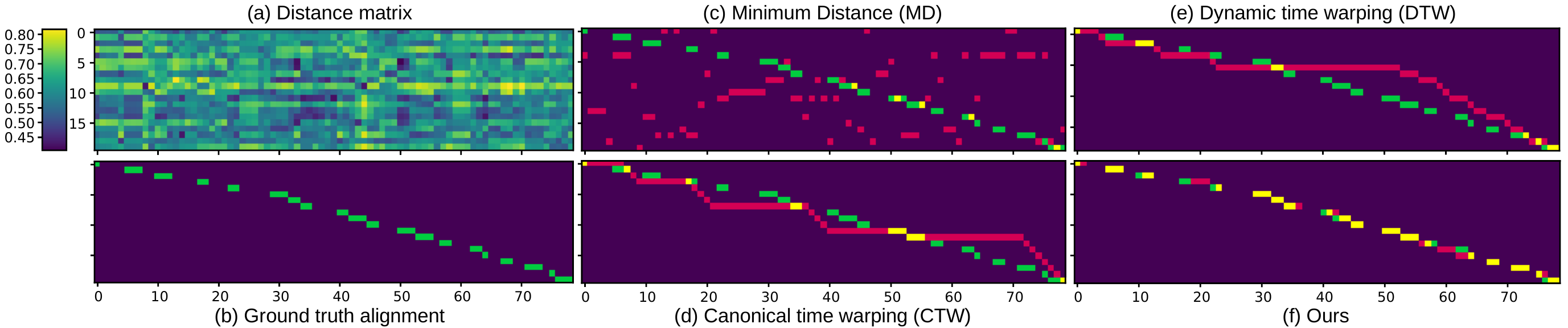}
	\label{fig:different-alignments}
	\vspace{-15pt}
	\caption{An alignment problem from HM-2 and the results. The vertical and horizontal axes represent the text sequence (sentences) and video sequence (clips) respectively. Green, red and yellow respectively represent the ground-truth alignment, the predicted alignment, and the intersection of two.}
	\label{fig:res_paths}
	\vspace{-12pt}
\end{figure*}

The MD method matches the most similar clip-sentence pairs which have the smallest distance compared to the others. We artificially boost this baseline using specific optimization for the two accuracy measures. For evaluation on video clips, we match every clip with the most similar sentence, but if the distance is greater than the threshold $0.7$, we consider the clip to be unmatched (i.e., a {\em null clip}). For sentence accuracy, we match every sentence with the most similar clip and do not assign {\em null} sentences. 

DTW computes the optimal path on the distance matrix. It uses the fact that the first sentence is always matched with the first clip, and the last sentence is always matched to the last clip, so the shortest path is between the upper left corner and lower right corner of the distance matrix. Note this is a constraint that NeuMATCH is not aware of. In order to handle null clips, we make use of the threshold again. In the case that one sentence is matched with several clips, the clips whose distances with the sentence are above the threshold will be assigned to null. We manually tuned the threshold to maximize the performance of all baselines. 
For CTW, we adopt source code provided in~\cite{ZhouD16} with the same assignment method as DTW.

\subsection{Results and Discussion}
Tables \ref{tab:results_action2} and \ref{tab:results_action3} show the performance under one-to-one and one-to-many scenarios, respectively.
On the one-to-one versions of the datasets HM-1 and HM-2, NeuMATCH demonstrates considerable improvements over the best baselines. It improves clip accuracy by 56.3 and 27.6 percentage points and improves sentence accuracy by 16.9 and 9.5 points. 
Unlike CTW and DTW, NeuMATCH does not have a major gap between clip and sentence performance. 

On the one-to-many versions of HM-1 and HM-2, as well as the YMS dataset, NeuMATCH again shows superior performance over the baselines. The advantage over the best baselines is 47.4, 19.7, and 1.7 points for clip accuracy, and 25.7, 1.1, and 2.9 for sentence IoU. Interestingly, NeuMATCH performs better on HM-1 than HM-2, but the other baselines are largely indifferent between the two datasets. This is likely due to NeuMATCH's ability to extract information from the matched stack. Since HM-1 is created by inserting random clips into the video sequence, the features of the inserted video clip match surrounding clips, but other aspects such as cinematography style may not match. This makes HM-1 easier for NeuMATCH because it can compare the inserted clip with those in the matched stack and detect style differences. It is worth noting that different cinematographic styles are commonly used to indicate memories, illusions, or imaginations. Being able to recognize such styles can be advantageous for understanding complex narrative content.

To further investigate NeuMATCH's performance without null clips, we additionally create a one-to-many dataset, HM-0, by randomly dividing every video clip into 1-to-5 smaller clips. Although NeuMATCH's advantage is reduced on HM-0, it's still substantial (12.5 points on both measures), showing that the performance gains are not solely due to the presence of null clips.

As we expect, the real-world YMS dataset is more difficult than HM-1 and HM-2. Still, we have a relative improvement of $17\%$ on clip accuracy and $39\%$ on sentence IoU over the closest DTW baseline. We find that NeuMATCH consistently surpasses conventional baselines across all experimental conditions. This clearly demonstrates NeuMATCH's ability to identify alignment from heterogenous video-text inputs that are challenging to understand computationally.

As a qualitative evaluation, Figure \ref{fig:res_paths} shows an alignment example. The ground alignment goes from the top left (the first sentence and the first clip) to the bottom right (the last sentence and the last clip). Dots in green, red, and yellow represent the ground truth alignment, the predicted alignment, and the intersection of the two, respectively. In the ground truth path (e), some columns does not have any dots because those clips are not matched to anything. As shown in (a), the distance matrix does not exhibit any clear alignment path. Therefore, MD, which uses only the distance matrix, performs poorly. 
The time warping baselines in (c) and (d) also notably deviate from the correct path, whereas NeuMATCH is able to recover most of the ground-truth alignment. For more alignment examples, we refer interested readers to the supplementary material. 

\subsection{Ablation Study}
In order to understand the benefits of the individual components of NeuMATCH, we perform an ablated study where we remove one or two LSTM stacks from the architecture. 
The model \textit{No Act\&Hist} lacks both the action stack and the matched stack in the alignment network. That is, it only has the text and the video stacks. The second model \textit{No Action} and the third model \textit{No History} removes the action stack and the matched stack, respectively. In the last model \textit{No Input LSTM}, we directly feed features of the video clip and the sentence at the tops of the respective stacks into the alignment network. That is, we do not consider the influence of future input elements.

Table~\ref{tab:ablations} shows the performance of four ablated models in the one-to-many setting. 
The four ablated models perform substantially worse than the complete model. This confirms our intuition that both the history and the future play important roles in sequence alignment. 
We conclude that all four LSTM stacks contribute to NeuMATCH's superior performance. 

\begin{table}
	\centering
	\renewcommand{\arraystretch}{0.8}
	\begin{tabular}{lcccc}
		\toprule
		& \multicolumn{2}{c}{\textbf{HM-1}} &   \multicolumn{2}{c}{\textbf{HM-2}} \\		
		\cmidrule(lr){2-3}
		\cmidrule(lr){4-5}
		& \textbf{clips} & \textbf{sent. IoU} &  \textbf{clips} & \textbf{sent. IoU} \\
		\midrule
		No Act\&Hist & 47.3 & 21.8 &  11.8 & 1.6 \\
		No Action & 49.9 & 23.0 &  29.6 & 16.1  \\
		No History & 57.6 & 33.4 &  28.3 & 17.0 \\
		No Input LSTMs & 54.8 & 24.6 &  27.9 & 8.3  \\
		\midrule
		\textbf{NeuMATCH} & \textbf{65.0} & \textbf{44.1} &  \textbf{37.7} & \textbf{20.0}  \\
		\bottomrule
	\end{tabular}
	\vspace{-5pt}
	\caption{Performance of ablated models in the one-to-many setting (3-action model).}
	\label{tab:ablations}
	\vspace{-10pt}
\end{table}

 \section{Conclusions}
In this paper, we propose NeuMATCH, an end-to-end neural architecture aimed at heterogeneous multi-sequence alignment, focusing on alignment of video and textural data. 
Alignment actions are implemented in our network as data moving operations between LSTM stacks. We show that this flexible architecture supports a variety of alignment tasks. Results on semi-synthetic and real-world datasets and multiple different settings illustrate superiority of this model over popular traditional approaches based on time warping. An ablation study demonstrates the benefits of using rich context when making alignment decisions.

 \PAR{Acknowledgement.}Pelin Dogan was partially funded by the SNF grant 200021\textunderscore153307/1.

\appendix

\section{Extensions to Multiple Sequences and Non-monotonicity}
\label{sec:extensions}

The basic action inventory tackles the alignment of two sequences. The alignment of more than two sequences simultaneously, like video, audio, and textual sequences, requires an extension of the action inventory. To this end, we introduce a parameterized \emph{Match-Retain} action. For three sequences, the parameters are a 3-bit binary vector where 1 indicate the top element from this sequence is being matched and 0 otherwise. Table \ref{tab:multi-seq-actions} shows one example using the parameterized Match-Retain. For instance, to match the top elements from Sequence A and B, the action is Match-Retain (110). The parameters are implemented as three separate binary predictions. 

The use of parameterized actions further enables non-monotonic matching between sequences. In all previous examples, matching only happens between the stack tops. Non-monotonic matching is equivalent to allowing stack top elements to match with any element on the matched stack. We propose a new parameterized action \emph{Match-With-History}, which has a single parameter $q$ that indicates position on the matched stack. To deal with the fact that the matched stack has a variable length, we adopt the indexing method from Pointer Networks~\cite{vinyals2015pointer}. The probability of choosing the $i^{\text{th}}$ matched element $r_i$ is 
\begin{subequations}
\begin{align}
P(q = i|\Psi_t) & = \frac{\exp (f(\psi_t, r_i))}{\sum_{j=0}^{L} \exp (f(\psi_t, r_j))}\\
f(\psi_t, r_i) & = v^\top \text{tanh} \left( W_{q} \begin{bmatrix} \psi_t \\ r_i \end{bmatrix} \right) 
\end{align}
\end{subequations}
where the matrix $W_{q}$ and vector $v$ are trainable parameters and $L$ is the length of the matched stack. 
\begin{table}[!b]
	\vspace{-7pt}
	\centering
	\begin{tabular}{lp{2.8em}p{2.7em}p{2.7em}p{4.7em}}
		\toprule
		& \textbf{Seq A} & \textbf{Seq B} &  \textbf{Seq C} & \textbf{Matched Stack} \\
		\midrule
		\textbf{Initial} & \textcircled{a}\circled{b}\textcircled{c}  & \circled{1}\circled{2}\circled{3} & \textcircled{x}\textcircled{y}\textcircled{z} & \\
		\midrule
		1. M-R(110) &  \textcircled{a}\circled{b}\textcircled{c}  & \circled{1}\circled{2}\circled{3} & \textcircled{x}\textcircled{y}\textcircled{z}  & [\textcircled{a}\circled{1}] \\
		2. Pop A & \circled{b}\textcircled{c} & \circled{1}\circled{2}\circled{3} & \textcircled{x}\textcircled{y}\textcircled{z} & [\textcircled{a}\circled{1}] \\
		3. Pop B & \circled{b}\textcircled{c} & \circled{2}\circled{3} & \textcircled{x}\textcircled{y}\textcircled{z} & [\textcircled{a}\circled{1}] \\
		4. M-R(011) & \circled{b}\textcircled{c} & \circled{2}\circled{3} & \textcircled{x}\textcircled{y}\textcircled{z} & [\circled{2}\textcircled{x}][\textcircled{a}\circled{1}] \\
		\bottomrule
	\end{tabular}
	\vspace{-5pt}
	\caption{An example action sequence for aligning three sequences.}
	\label{tab:multi-seq-actions}
	\vspace{-7pt}
\end{table}

\clearpage
\balance
{\small
\bibliographystyle{ieee}
\bibliography{egbib}
}

\newpage
\onecolumn
\begin{mytitlepage}
\def\@thanks{}
\title{\textbf{A Neural Multi-sequence Alignment TeCHnique (NeuMATCH): Supplemental Material}}
\maketitle

In this supplementary material, we first give details on the segmentation of videos into clips. Next, we show more alignment results computed by our approach on the datasets HM-1, HM-2, and YMS that require one-to-many matching and contain clips that do not match any sentences (i.e., \textit{null} clips). For illustration purposes, each figure below represents only a small portion (6-12 consecutive clips) of the entire aligned sequence. Each frame represents a video clip. The aligned sentences are shown with wide brackets below or above the clips.  
\\
\section{Implementation Details}
As discussed in Sec. 3.2 in the main paper, we customize the action inventory using knowledge of the dataset. For one-to-one matching with {\em null} video clips, we use the actions Pop Clip and Match. For one-to-many matching with {\em null} video clips, we use Pop Clip, Pop Sentence, and Match-Retain Sentence. For all the experiments, action decoding is done greedily.

For the joint pre-training, we use 500 dimensions for the LSTM sentence encoder and 300 for the joint embeddings. The dimensions of the word and image embedding are 300 and 4096, respectively, while the margin in the ranking objective function is $\alpha=0.05$. $L_2$ regularization is used to prevent over-fitting. The batch size is set to $32$ and the number of contrastive samples is 31 for every positive pair. The model is trained with the Adam optimizer using a learning rate of $10^{-4}$ and gradient clipping of 2.0. Early stopping on the validation set is used to avoid over-fitting. 

The alignment network uses 300 dimensions for the video and text stacks, 20 dimensions for the matched stack and 8 for the history stack. Optionally, we feed two additional variables into the fully connected layer: the numbers of elements left in the video and text stacks to improve the performance on very long sequences in the YMS dataset. 
The alignment network is first trained with the encoding networks fixed with a learning rate of $0.001$. After that, the entire model is trained end-to-end with a learning rate of $10^{-5}$. For HM-0, HM-1, and HM-2, we use the original data split of LSMDC. For YMS, we use a 80/10/10 split for training, validation and test sets. 

\PAR{Details of Video Segmentation}
The video segmentation can be achieved using any shot boundary detection algorithm. In this work, we segment the input videos into video clips by a Python/OpenCV-based scene detection program\footnote{https://github.com/Breakthrough/PySceneDetect} that uses threshold/content on a given video. For the parameters, we choose the \textit{content-aware} detection method with the \textit{threshold} of $20$ and \textit{minimum length} of $5$ frames. Having a low threshold and minimum length usually results in over-segmentation. However, NeuMATCH can handle this resulting over-segmentation with the ability of one-to-many matching.

\section{Alignment Results}
\subsection{Successful results for Hollywood Movies 1 (HM-1)}
The video sequences in HM-1 contain clips from other movies that are inserted into the original sequence, as explained in the main paper. 
\\

\begin{figure*}[h!]
	\centering
	\includegraphics[width=\textwidth]{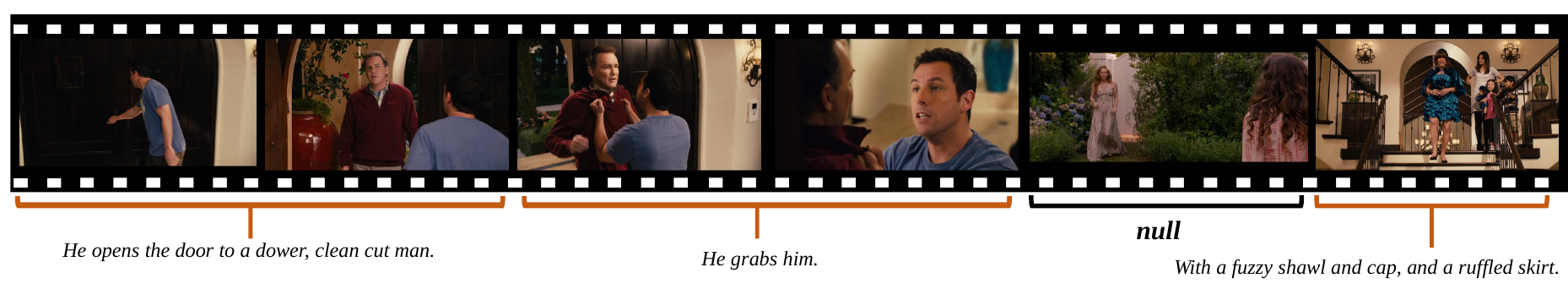}
	\label{fig:hm1_jack_and_jill}
	\vspace{-10pt}
	\caption{From the movie \textit{Jack and Jill} in dataset HM-1. The fifth frame is from the movie \textit{This is 40}, which is successfully assigned as \textit{null}. Note the last two frames have very similar content (two women in dresses) to the sentence \textit{``With a fuzzy shawl and cap, and a ruffled skirt.''}, but our algorithm was able to identify them correctly. }
\end{figure*}

\begin{figure*}[h!]
	\centering
	\includegraphics[width=\textwidth]{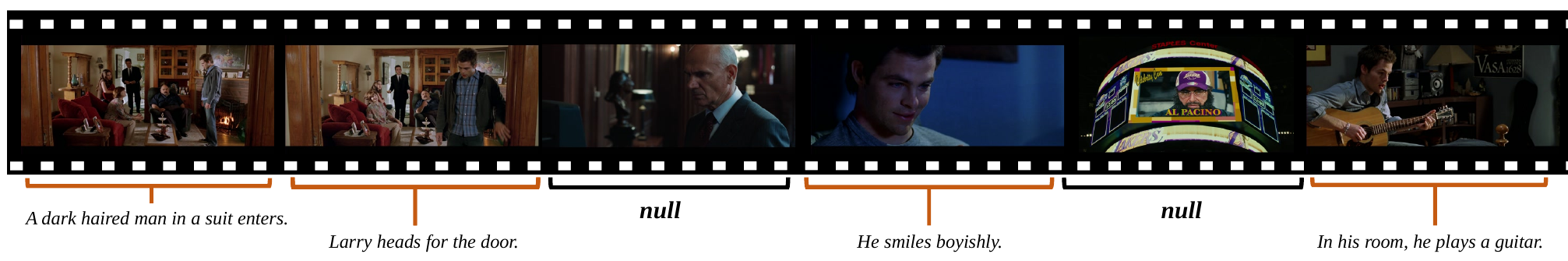}
	\label{fig:hm1_blind_dating}
	\vspace{-10pt}
	\caption{From the movie \textit{Blind Dating} in dataset HM-1. The third frame is from the movie \textit{Inside Man}, and the fifth frame is from the movie \textit{Jack and Jill}, which are correctly assigned to \textit{null}.}
\end{figure*}

\begin{figure*}[h!]
	\centering
	\includegraphics[width=\textwidth]{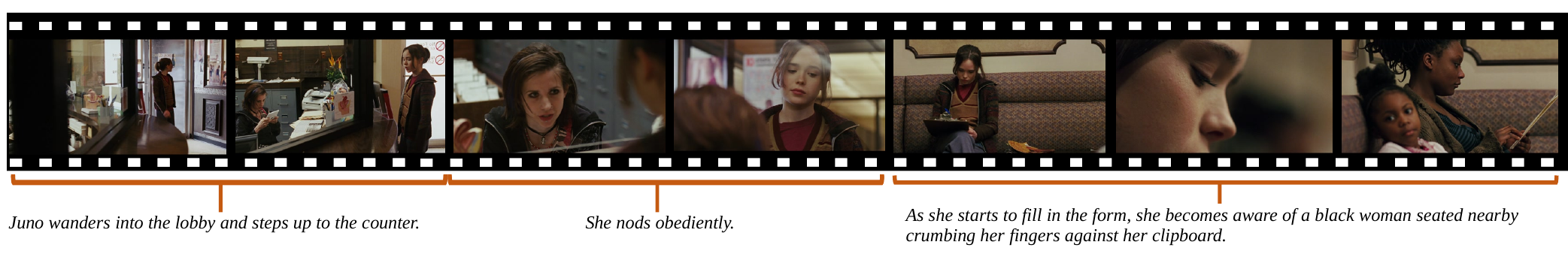}
	\label{fig:hm2_juno}
	\vspace{-10pt}
	\caption{From the movie \textit{Juno} in dataset HM-1. The one-to-many assignment for the last three clips is correctly identified even when there is a significant perspective and content change through the clips.}
\end{figure*}

\newpage

\subsection{Successful results for Hollywood Movies 2 (HM-2)}

Each video sequence in HM-2 consists of consecutive clips from a single movie, where some sentences were discarded in order to create \textit{null} clips. It still requires one-to-many matching of the sentences and the assignment of \textit{null} clips. 

\begin{figure*}[h!]
	\centering
	\includegraphics[width=\textwidth]{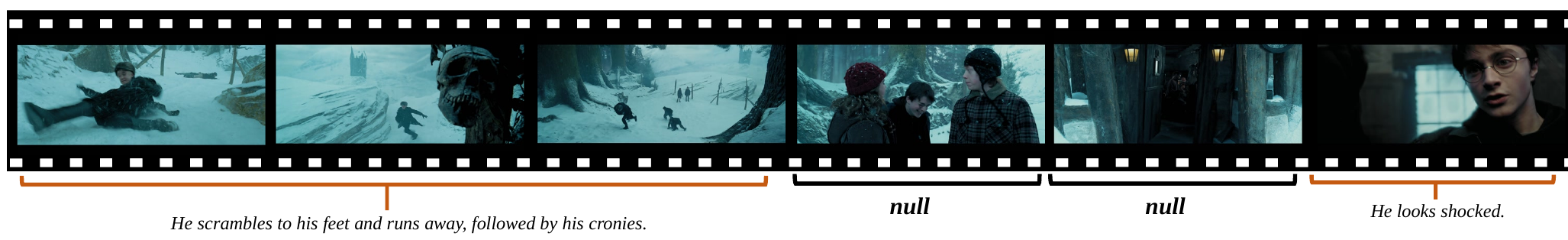}
	\label{fig:hm2_azkaban}
	\vspace{-10pt}
	\caption{From the movie \textit{Harry Potter and the Prisoner of Azkaban} in dataset HM-2}
\end{figure*}

\begin{figure*}[h!]
	\centering
	\includegraphics[width=\textwidth]{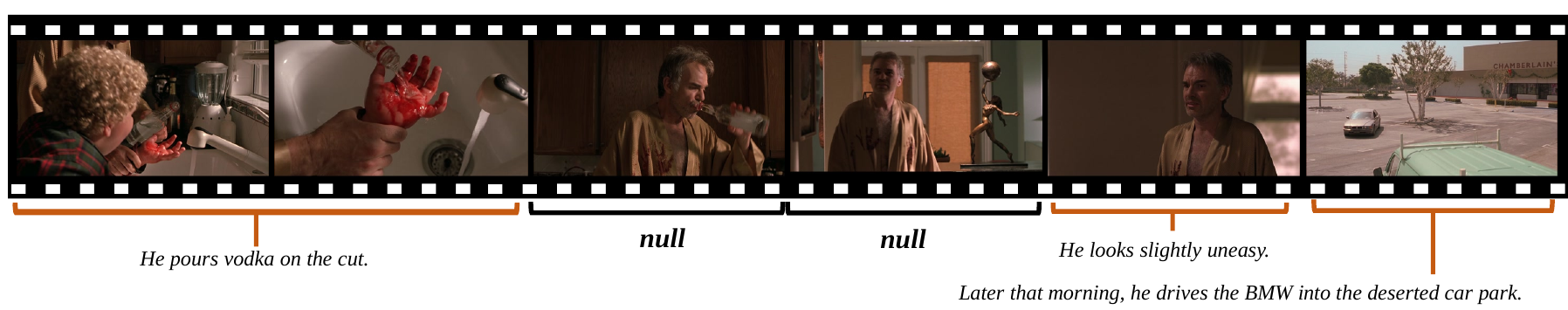}
	\label{fig:hm2_bad_santa}
	\vspace{-10pt}
	\caption{From the movie \textit{Bad Santa} in dataset HM-2}
\end{figure*}

\begin{figure*}[h!]
	\centering
	\includegraphics[width=\textwidth]{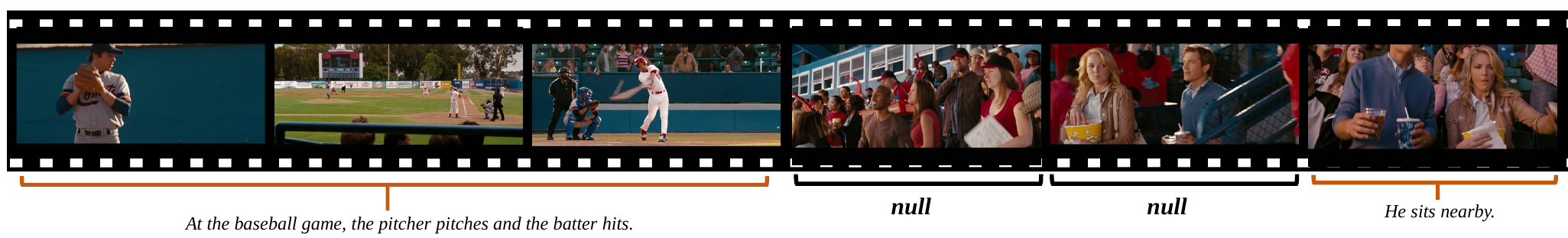}
	\label{fig:hm2_ugly_truth}
	\vspace{-10pt}
	\caption{From the movie \textit{The Ugly Truth} in dataset HM-2. The third clip contains a vodka bottle, which is mentioned in first sentence. The fourth and the fifth clips are very similar. However, the algorithm finds the correct alignment.}
\end{figure*}

\begin{figure*}[h!]
	\centering
	\includegraphics[width=\textwidth]{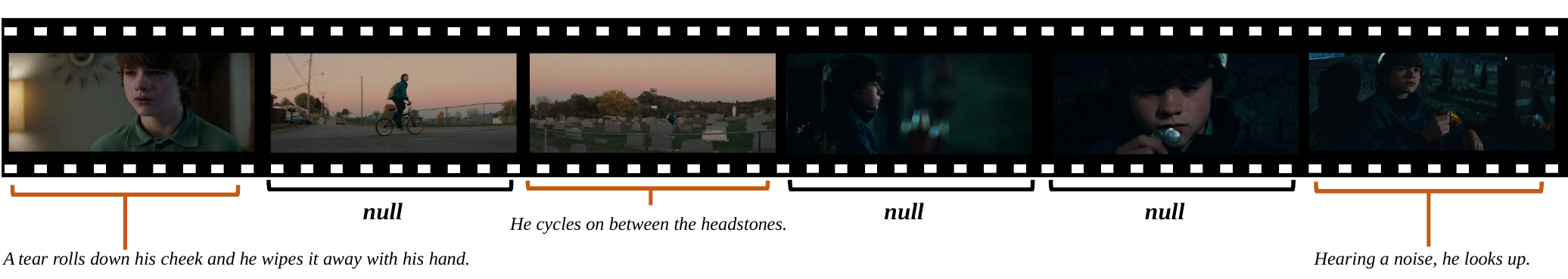}
	\label{fig:hm2_super_8}
	\vspace{-10pt}
	\caption{From the movie \textit{Super 8} in dataset HM-2. The boy and the bicycle are visible in both the second and the third clips, but the headstones only appear in the third clip. The algorithm makes the correct decision. }
\end{figure*}

\begin{figure*}[h!]
	\centering
	\includegraphics[width=\textwidth]{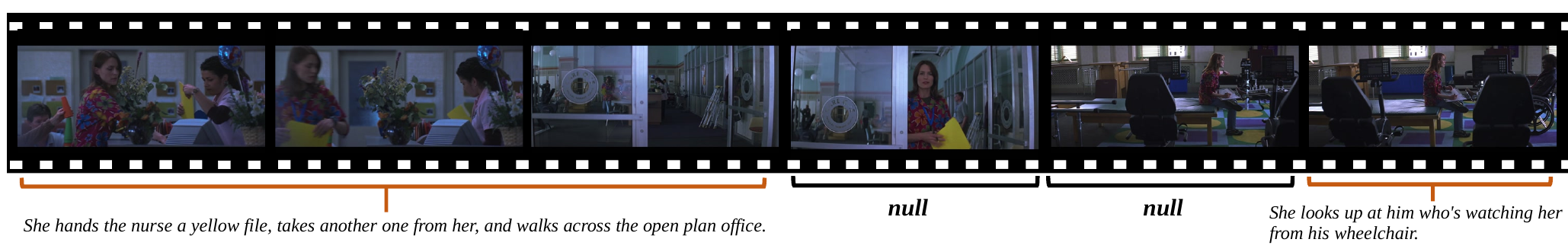}
	\label{fig:hm2_unbreakable}
	\vspace{-10pt}
	\caption{From the movie \textit{Unbreakable} in dataset HM-2. The wheelchair is only visible in the last clip and the algorithm successfully picks that up. }
\end{figure*}

\newpage

\subsection{Successful results for YouTube Movie Summaries (YMS)}

In the YMS dataset, the sentences are longer than HM-1 and HM-2, and they tend to describe multiple events. We asked the annotators to break them down into small units, which allows them to precisely align the text with the video sequence. These sequences tend to be much more complex than HM-1 and HM-2. 

\begin{figure*}[h!]
	\centering
	\includegraphics[width=\textwidth]{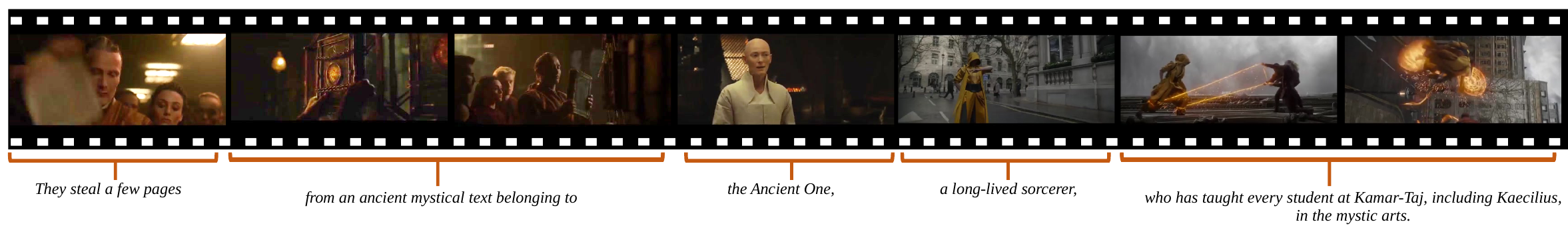}
	\label{fig:yms_doctor}
	\vspace{-10pt}
	\caption{From the movie \textit{Doctor Strange} in dataset YMS. The original video is available at \url{https://www.youtube.com/watch?v=fZeW-KUXHKY}}
\end{figure*}

\begin{figure*}[h!]
	\centering
	\includegraphics[width=\textwidth]{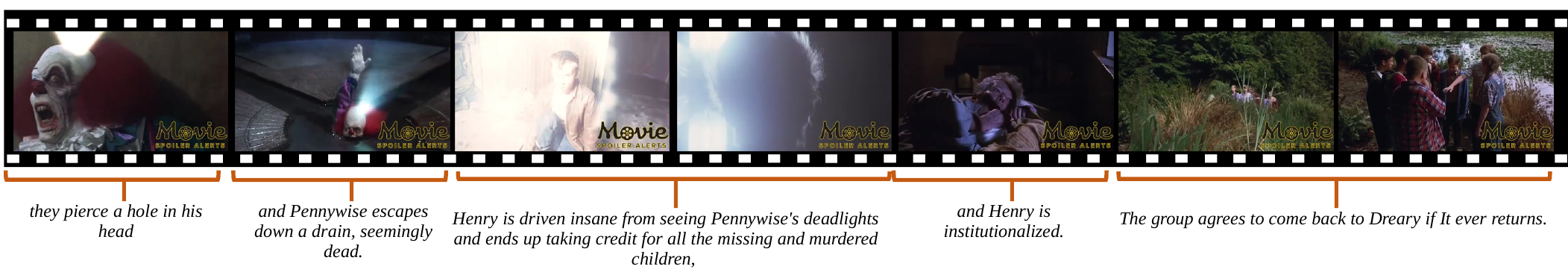}
	\label{fig:it}
	\vspace{-10pt}
	\caption{From the movie \textit{It (1990)} in dataset YMS. The original video is available at \url{https://www.youtube.com/watch?v=c-sIoODkpuU}}
\end{figure*}
 
\begin{figure*}[h!]
	\centering
	\includegraphics[width=\textwidth]{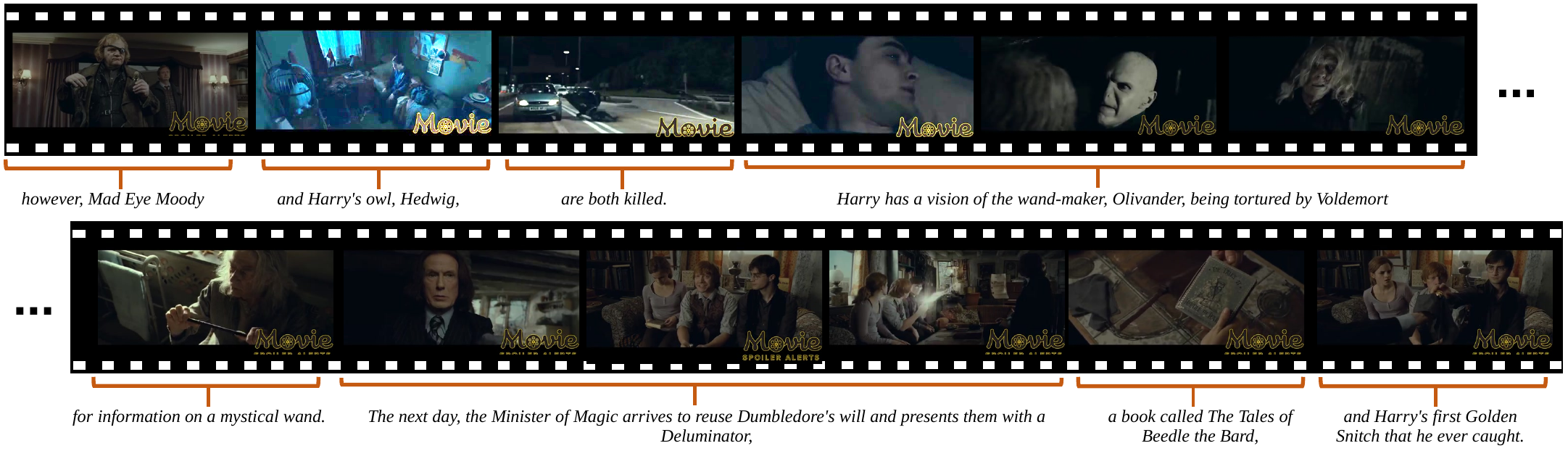}
	\label{fig:yms_deathly}
	\vspace{-10pt}
	\caption{From the movie \textit{Harry Potter and the Deathly Hallows} in dataset YMS. The original video is available at \url{https://www.youtube.com/watch?v=nfuRErj9TkY}}
\end{figure*}

\subsection{Failure Cases}

We present two failure cases below. The ground truth is shown with green brackets and NeuMATCH's predictions are with orange brackets. 

\begin{figure*}[h!]
	\centering
	\includegraphics[width=\textwidth]{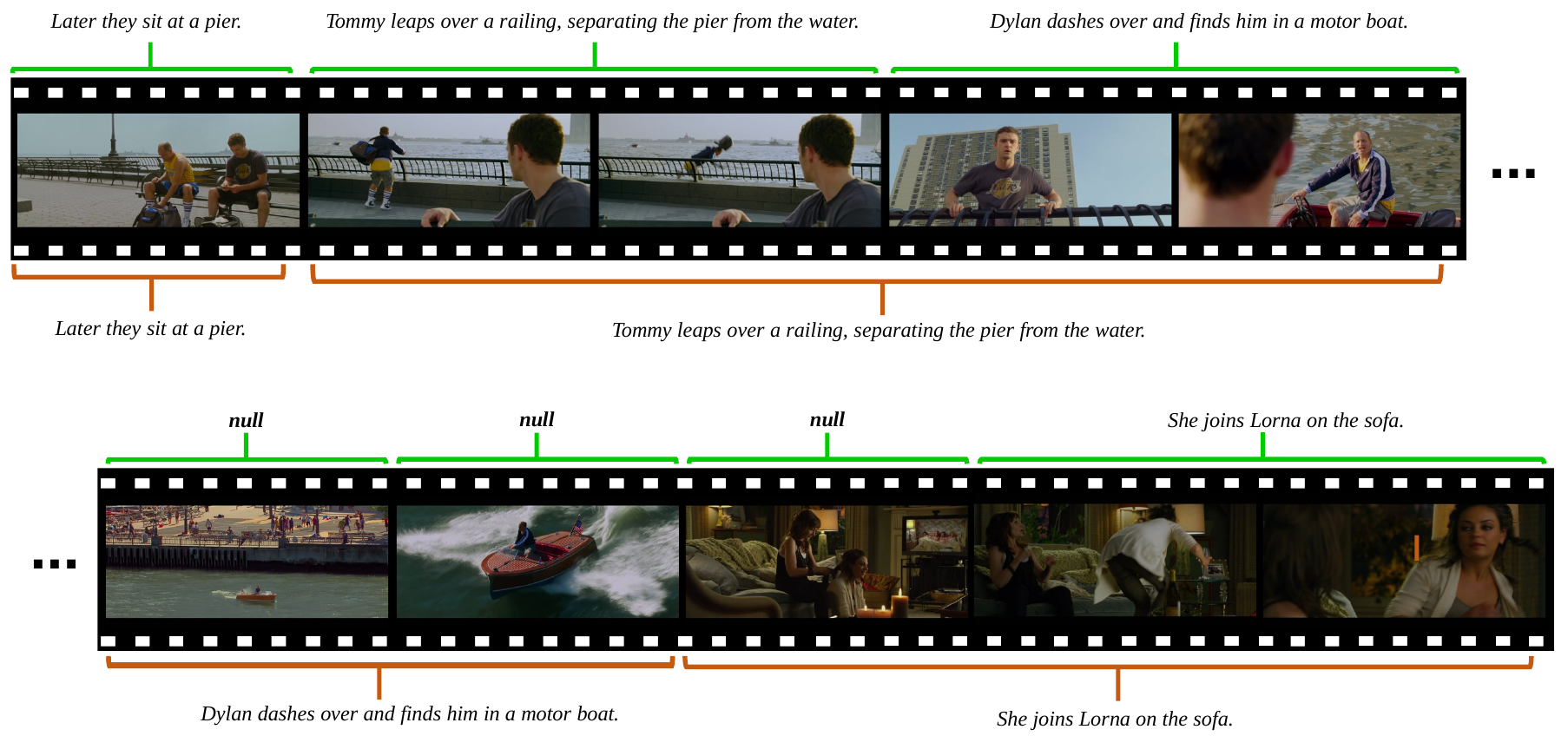}
	\label{fig:fail_friends}
	\vspace{-10pt}
	\caption{From the movie \textit{Friends with Benefits} in dataset HM-2. The assignments with green marks represent the ground truth while orange marks represent our result. The first failure is that the second sentence is matched with two more clips, but the additional clips also contain the ``railing'' and the ``water'', which may have confused the algorithm. Similarly, the boat appears in the sixth and seventh clips, which may have caused the wrong alignment with the third sentence. }
\end{figure*}

\makeatletter
\setlength{\@fptop}{0pt}
\makeatother
\begin{figure*}[h!]
	\centering
	\includegraphics[width=\textwidth]{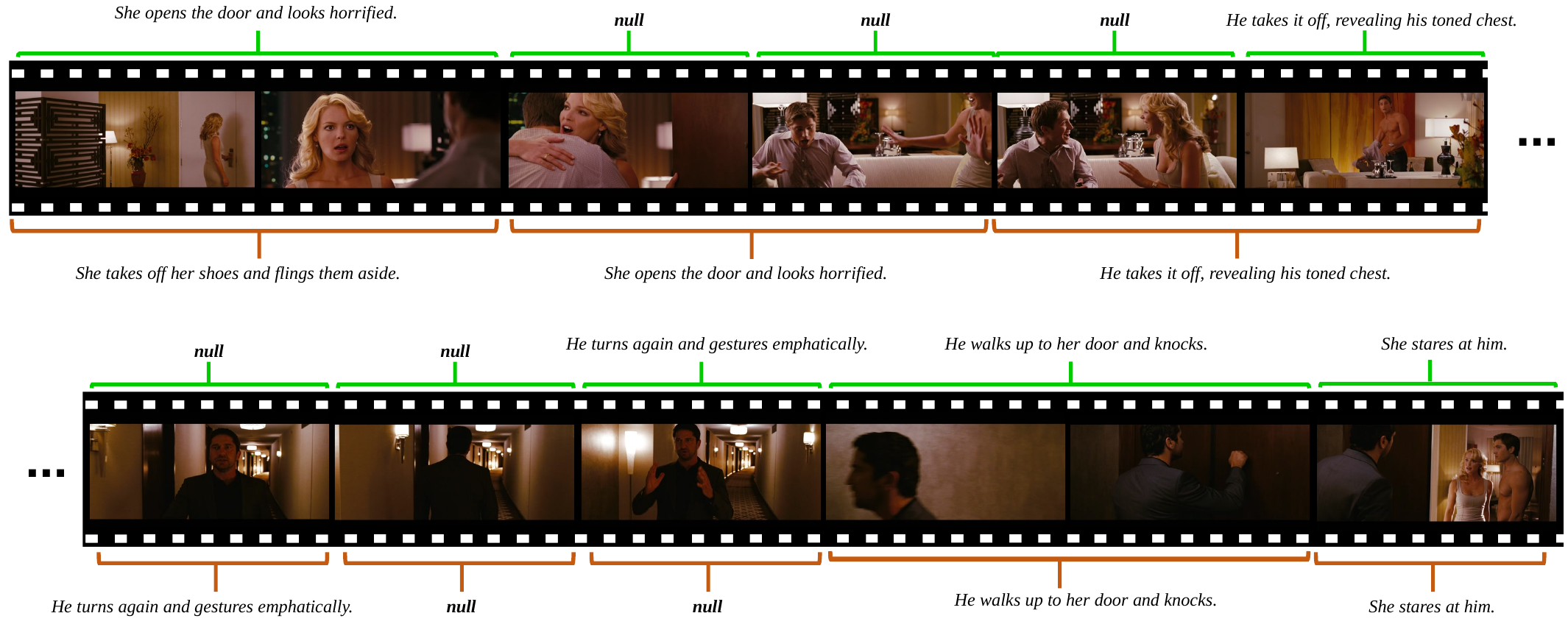}
	\label{fig:fail_ugly}
	\vspace{-10pt}
	\caption{From the movie \textit{The Ugly Truth} in dataset HM-2. The assignments with green marks represent the ground truth while orange marks represent our result. }
\end{figure*}

 \end{mytitlepage}

\end{document}